\documentclass[10pt,twocolumn,letterpaper]{article}

\usepackage[pagenumbers]{cvpr} % To force page numbers, e.g. for an arXiv version

\definecolor{cvprblue}{rgb}{0.21,0.49,0.74}

\usepackage[accsupp]{axessibility}  % Improves PDF readability for those with disabilities.
\usepackage{float}
\usepackage{lipsum}
\usepackage{multirow}
\usepackage{multicol}
\usepackage{booktabs} 
\usepackage{arydshln}
\usepackage{xcolor}
\usepackage[pagebackref,breaklinks,colorlinks,allcolors=cvprblue]{hyperref}

\def\paperID{14803}
\def\confName{CVPR}
\def\confYear{2025}

\title{MixerMDM: Learnable Composition of Human Motion Diffusion Models}

\author{
Pablo Ruiz-Ponce$^{
1}$, German Barquero$^{2}$, Cristina Palmero$^{3}$, Sergio Escalera$^{2}$, José García-Rodríguez$^{1}$ \\
\normalsize
$^{1}$Universidad de Alicante, Spain\\
\normalsize
$^{2}$Universitat de Barcelona and Computer Vision Center, Spain\\
\normalsize
$^{3}$King's College London, UK\\
\normalsize
{\tt pruiz@dtic.ua.es} \\
\normalsize
\url{https://pabloruizponce.com/papers/MixerMDM}
}

\begin{document}

\twocolumn[{
\renewcommand\twocolumn[1][]{#1}
\maketitle
\begin{center}
    \centering
    \vspace{-0.81cm}\includegraphics[width=0.95\textwidth]{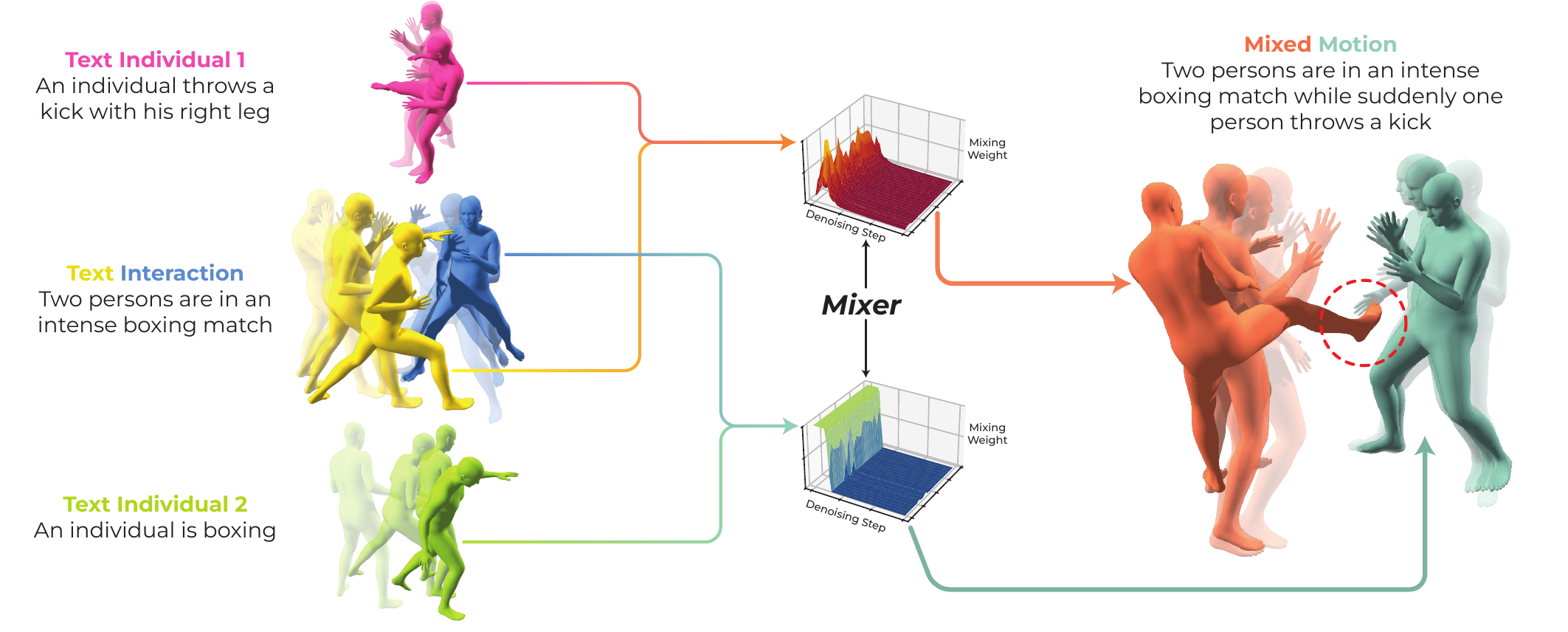}
        \vspace{-0.5cm}
    \captionof{figure}{We introduce \textbf{MixerMDM}, the first learnable model composition technique for combining pre-trained text-conditioned human motion diffusion models. MixerMDM has demonstrated a consistent ability to generate highly controllable human interactions by combining a model that generates individual motions from textual descriptions with a model that creates human-human interactions.}
    \label{fig:teaser}
\end{center}
}]

\begin{abstract}
Generating human motion guided by conditions such as textual descriptions is challenging due to the need for datasets with pairs of high-quality motion and their corresponding conditions. The difficulty increases when aiming for finer control in the generation. To that end, prior works have proposed to combine several motion diffusion models pre-trained on datasets with different types of conditions, thus allowing control with multiple conditions. However, the proposed merging strategies overlook that the optimal way to combine the generation processes might depend on the particularities of each pre-trained generative model and also the specific textual descriptions. In this context, we introduce MixerMDM, the first learnable model composition technique for combining pre-trained text-conditioned human motion diffusion models. Unlike previous approaches, MixerMDM provides a dynamic mixing strategy that is trained in an adversarial fashion to learn to combine the denoising process of each model depending on the set of conditions driving the generation. By using MixerMDM to combine single- and multi-person motion diffusion models, we achieve fine-grained control on the dynamics of every person individually, and also on the overall interaction. Furthermore, we propose a new evaluation technique that, for the first time in this task, measures the interaction and individual quality by computing the alignment between the mixed generated motions and their conditions as well as the capabilities of MixerMDM to adapt the mixing throughout the denoising process depending on the motions to mix.
\end{abstract}    
\vspace{-0.6cm}
\section{Introduction}
\label{sec:intro}
Creating synthetic human motion that closely mimics real individuals is important for applications such as animation~\cite{starke2024categorical}, virtual reality~\cite{ma2021pixel, park2022metaverse}, and robotics \cite{suzuki2022augmented}. Human motion generation techniques have experienced remarkable progress in recent years~\cite{related:hmg-survey}.
However, unlike other generative fields such as image or text where data is abundant, high-quality human motion data remains scarce and costly to obtain and annotate. Currently, there exists a variety of small, specialized datasets that focus on specific ranges of motion over the entire human motion manifold (e.g., \cite{data:humanml3d, data:BABEL}). However, there is no consensus on the data representation format (positions/rotations, relative/global, redundancy, etc.), and the conditions to generate the motions vary widely (text, audio, other motions, scene, etc.). This heterogeneity makes it difficult to integrate such datasets to expand the distribution of possible human motions. While recent attempts have been made to create generalist models for human motion generation~\cite{lmm}, the general trend has been the development of highly specialized models tailored to the characteristics of individual datasets.

To exploit the specific capabilities of these models into new motions, we introduce MixerMDM, a model composition technique that learns how to mix motions generated by specialized motion models without re-training. Mixing the distinct motions generated by these models can result in the generation of a unique new motion preserving some of their specialized generative capabilities, thereby expanding the coverage of possible motions. MixerMDM is, to the best of our knowledge, the first learnable approach that dynamically mixes text-conditioned human motion diffusion models, leveraging each model's unique strengths to enhance task-specific performance. Unlike previous methods~\cite{related:interaction-1, in2in}, MixerMDM performs a unique mixing strategy using the mixing weights predicted by the \textit{Mixer} depending on the motions generated by two pre-trained models, the conditions used to generate them, and the step in the denoising chain. Furthermore, we propose different modalities of mixing, giving MixerMDM the capability of blending motions at global,  motion duration, body joint, or spatio-temporal (duration$*$joints) levels. As an application of our approach, in this paper we focus on the blending of single-person (i.e. individual) motions and human-human interactions generated by text-conditioned human motion diffusion models. This results in the consistent generation of unique new interactions leveraging intrapersonal capabilities of the individual model -such as diverse and precise conditioning of individual motions- with interpersonal capabilities of the interaction model -such as global positioning and orientation in multi-human scenarios. 

To train MixerMDM, we introduce a pipeline based on adversarial losses as in Generative Adversarial Networks (GANs)~\cite{goodfellow2014generative} to take advantage of the prior knowledge embedded in the pre-trained models. The only components that undergo training are the \textit{Mixer} module, which determines how to best blend the motions, and a discriminator per model, that captures the particularities of each model. Since no ground truth exists for the combination of different motions, we utilize the predictions generated by the pre-trained models as positive samples for their respective discriminators. In addition, the \textit{Mixer} uses directly the final predictions from the pre-trained models to mix the motions, allowing it to adapt directly to the data distribution without relying on the particularities of each model architecture. This flexibility allows seamless swapping of models trained with the same dataset without additional training.

Finally, we introduce an evaluation procedure to assess the mixing quality. To do so, we evaluate the alignment of the generated mixed motion and the condition used to generate it with respect to each of the distributions of the pre-trained models used in the combination. We further measure the capabilities of our proposal to adapt the mixing depending on the specific motions, conditions, and denoising timestep to blend. 

The main contributions of this paper are as follows:

\begin{itemize}
\item We propose MixerMDM, a method for dynamically mixing distinct text-conditioned human motion diffusion models while keeping their specific capabilities. As a result, our approach consistently mixes individual motions and human-human interactions, achieving more fine-grained individual controllability than previous methods.
\item We introduce an adversarial training strategy using multiple discriminators. This approach improves the ability of our method to learn robust blendings as the pre-trained model outputs are used as ground truth in the absence of a real one. Additionally, as the motions predicted are directly used as inputs to the \textit{Mixer}, it allows to swap the pre-trained models without additional training.
\item Given the lack of quantitative metrics for the task of mixing motions generated by different models, we propose a new evaluation pipeline to assess the quality of the mixing as well as the capabilities of the model to dynamically adapt the mixing strategy during the denoising process.
\end{itemize}
\section{Related Work}
\label{sec:related}

\noindent\textbf{Text-Conditioned Human Motion Generation.} A recent review \cite{related:hmg-survey} has highlighted significant advances in human motion generation. Early works focused on aligning text and motion latent spaces using Kullback-Leibler divergence loss~\cite{related:t2m,related:language2pose,related:t2m-motionclip,related:t2m-temos}, but these methods often suffer from latent space misalignments and semantic mismatches due to limited data. Recent successes in autoregressive models, such as Large Language Models \cite{related:bert, related:llama, related:llm-survey} powered by Transformers \cite{related:transformers}, have inspired new motion generation methods~\cite{related:t2m-tm2t,related:t2m-t2mgpt,related:t2m-attt2m,related:t2m-motiongpt,wan2024tlcontrol}. These methods tokenize motions into discrete codes and use Transformers to convert text tokens into motion tokens. However, tokenizing motion is complex, and autoregressive models cannot capture bidirectional dependencies. Methods like MMM \cite{related:t2m-mmm} and MoMask \cite{related:momask} address this using masked attention. Diffusion models \cite{related:diffusion-1, related:diffusion-2} have become popular for generative tasks \cite{related:diffusion-survey,dai2024motionlcm,zhou2024emdm}, achieving state-of-the-art results in text-to-motion generation. FLAME \cite{related:t2m-flame} and MotionDiffusion \cite{related:t2m-motiondiffuse} use traditional diffusion models with Transformers as noise predictors. MDM \cite{classes-3} employs $x_0$ reparametrization \cite{xiao2021tackling, barquero2023belfusion}, allowing kinematic loss functions to improve motion quality. Other approaches incorporate physical constraints \cite{related:t2m-fisica}, use latent diffusion for faster sampling \cite{related:mld}, or apply retrieval techniques \cite{related:remodiffuse,fujiwara2024chronologically}. Despite slow inference, diffusion models produce realistic and diverse samples \cite{related:diffusion-diversity}. 

\noindent\textbf{Text-Conditioned Human-Human Interaction Generation.} In addition to single human motion generation, several recent works focus on the generation of human-human interactions using textual descriptions. ComMDM \cite{related:interaction-1} extends MDM to multi-human interactions using a cross-attention module. Similarly, \cite{related:interaction-5} uses a shared cross-attention module to connect two models for asymmetric interactions. InterGen~\cite{data:intergen} proposes the use of cooperative denoisers in the recently introduced InterHuman dataset. MoMat-MoGen \cite{related:momatmogen} extends the retrieval diffusion model from \cite{related:remodiffuse} for human interactions. More recently, methods such as in2IN \cite{in2in} and InterMask \cite{intermask} propose further improvements to enhance the generation of human interactions, achieving state-of-the-art performance.

\noindent\textbf{Human Motion Composition.} Mixing data of the same modality has a long history in fields like image synthesis \cite{related:mixing:images1,related:mixing:images2,related:mixing:images3,related:mixing:images4}, where new data can be generated that retains characteristics from multiple sources. The iterative nature of diffusion models allows for a smoother mixing of such data \cite{multidiffusion,diffcollage,collaborative,liew2022magicmix,liu2022compositional}. In the context of human motion, composition can be broadly categorized into temporal and spatial approaches. Temporal composition focuses on combining individual motion sequences into a larger cohesive sequence \cite{related:teach, related:interaction-1, related:flowmdm}, enabling smooth and realistic transitions between different actions. By contrast, spatial composition merges multiple motions to create a new motion of the same length, incorporating specific elements such as actions, trajectories, or joint movements from the original motions~\cite{related:t2m-motionclip,related:sinc}. More generally, \cite{related:interaction-1} introduced a model composition technique to combine the sampling processes of two different diffusion models, generating a blended motion through fixed-weight mixing during the denoising process. Building on this, in2IN \cite{in2in} proposed a weight scheduler to adjust the mixing throughout the denoising process, resulting in more versatile and refined outputs. However, both methods require the mixing weight or scheduler to be predefined manually prior to inference. In contrast, and inspired by \cite{collaborative}, we introduce the \textit{Mixer} module capable of learning a unique mixing weight for each motion at each timestep of the denoising process. While this approach incurs a minor additional computational cost, it provides an adaptable mixing that effectively preserves features from both motions being combined.

\noindent\textbf{Adversarial Training.} GANs~\cite{goodfellow2014generative} revolutionized the generative field when first introduced, employing a generator and a discriminator in a competitive framework where the generator aims to fool the discriminator, while the discriminator is trained to distinguish real from generated data. This training paradigm deviates from typical loss functions, which directly predict the ground truth. Despite numerous advancements on this initial idea \cite{gan:survey}, GAN-based architectures have been largely supplanted by diffusion models \cite{gan:diffusionbeat}. However, the adversarial training paradigm is still applied to enhance diffusion models \cite{gan:diffusion1,gan:diffusion2,gan:diffusion3,gan:diffusion4,gan:diffusion5}, providing guidance and improving generation quality. In our approach, we leverage this paradigm to train the module responsible for determining how two motions are mixed. Specifically, we introduce a discriminator for each pre-trained model, compelling the generator to deceive both discriminators and thus achieving a mixing strategy that preserves the core characteristics from each model.
\section{Method}
\label{sec:method}

\begin{figure*}[ht]
  \centering
  \includegraphics[width=\textwidth]{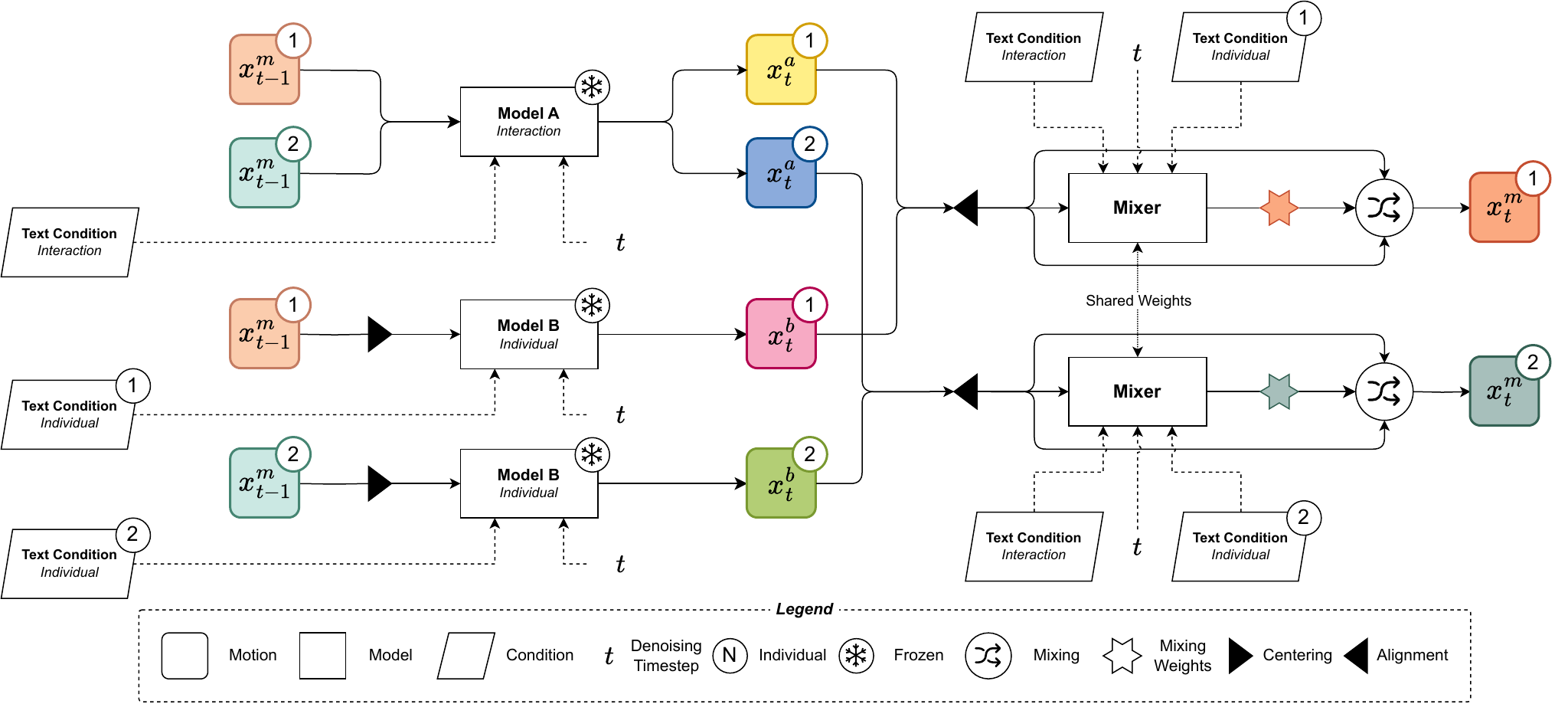}
  \caption{\textbf{MixerMDM pipeline.}  At each timestep $t$ of the denoising process, a mixed motion is generated by first obtaining motions from separate text-conditioned pre-trained motion diffusion models. Using these motions and their conditions, the Mixer predicts unique mixing weights that are subsequently used in the Mixing procedure to blend the generated motions and obtain the mixed motion $x^m_t$.}
  \label{fig:architecture}
  \vspace{-3mm}
\end{figure*}

The goal of MixerMDM is to generate a new motion sequence by combining, at each step of the denoising process, the motions generated by two models previously trained in a specialized dataset (i.e., pre-trained models). MixerMDM englobes the whole pipeline (\cref{sec:method:pipeline}, \cref{fig:architecture}) of generating the motions from the pre-trained models, determining how these particular motions are going to be mixed for this specific step of the denoising chain (\cref{sec:method:mixer}), and performing the mixing (\cref{sec:method:mixing}). We train the whole procedure using adversarial losses to use the specific knowledge of each pre-trained model as pseudo-ground truth (\cref{sec:method:discriminative}). 

Building on the work of \cite{in2in}, this paper focuses on mixing an interaction model with a specialized individual model. We consider interaction and individual models as text-conditioned diffusion models that generate human-human interactions and single-human motions from textual descriptions, respectively. The dynamics of human-human interactions can be divided into two distinct levels: intrapersonal and interpersonal. The former refers to the specific movements of an individual in isolation, whereas the latter encompasses not only the motion but also the overall trajectory and orientation of the body with respect to the other individual involved in the interaction. The goal is to combine the interpersonal capabilities of the interaction model with the intrapersonal capabilities of the individual model to be able to generate new diverse interactions with fine-grained control of the specific motions of the individuals. Nevertheless, given the characteristics of the method, it could be applied to mix any type of motion diffusion model.

\subsection{Pipeline} 
\label{sec:method:pipeline}
The mixed motion $x^m_t$ at timestep $t$ of the denoising chain is the result of combining two motion sequences $\{x^a_t,x^b_t\}$ generated by two distinct pre-trained models $\{\mathcal{M}^a,\mathcal{M}^b\}$. These two motions will be used by the \textit{Mixer} (\cref{sec:method:mixer}) to predict the mixing weight $w_t$  used by the \textit{Mixing} procedure to blend both motions. \cref{fig:architecture} depicts this process for the specific case of mixing interaction and individual motions.

The first step to adapt this general pipeline to the specific case of combining human-human interactions generated by $\mathcal{M}^a$ with individual motions generated by $\mathcal{M}^b$ is the duplication of $\mathcal{M}^b$ to generate two individual motions instead of one. Additionally, as stated in \cite{in2in}, whereas individual models focus on intrapersonal dynamics, interaction models place greater emphasis on interpersonal dynamics. The pre-trained models expect as input the denoised motion from the previous step. However, this motion has to be somewhere within the learned motion manifold of this model. Since in MixerMDM the inputs of the pre-trained models are the mixed motion from the previous steps, some transformations 
are required to match the input format of each of the models. In particular, the \textit{centering} function canonicalizes a motion's initial global translation and orientation. Conversely, the \textit{alignment} function undoes the \textit{centering} by harmonizing the global positions and trajectories of two motions. It is applied to the the individual model with the trajectory and orientation of its respective individual generated by the interaction model (see sup. material).

\subsection{Mixer}
\label{sec:method:mixer}
\begin{figure}[h]
    \centering
    \includegraphics[width=\linewidth]{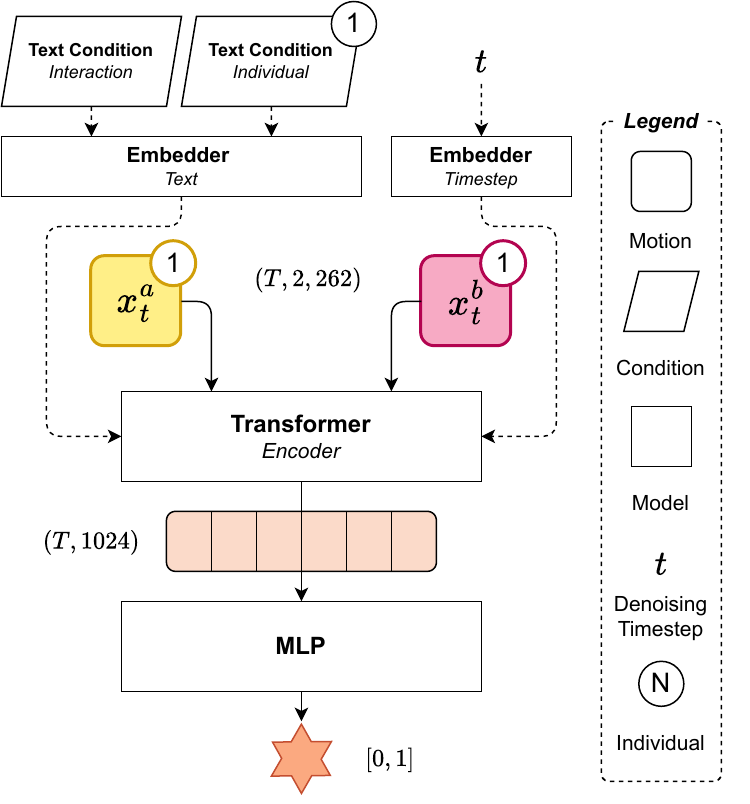}
    \caption{\textbf{Mixer architecture.} The \textit{Mixer} is composed of a Transformer encoder that takes as input both generated motions by the pre-trained models, their respective conditions, and the actual timestep of the denoising process. This encoder generates a latent representation, which is decoded by an MLP that outputs the mixing weights. $T$: number of frames of the motion sequence.}
    \label{fig:mixer}
    \vspace{-2mm}
\end{figure}

The \textit{Mixer} (see \cref{fig:mixer}) is a lightweight and specialized module capable of learning to combine two motions from different models. This module receives as input the two motions from the pre-trained models, the actual timestep $t$ of the denoising process, and the conditions employed by the two models to generate the motions $\{c^a, c^b\}$. These inputs will be processed by a Transformer~\cite{related:transformers} encoder that will transform them into a high-dimensional representation that then will be processed by a Multi-Layer Perceptron (MLP) to output a mixing weight $w_t$. This $w_t$ will be used to blend $x^a_t$ and $x^b_t$ following the process described in \cref{sec:method:mixing}. $w_t$ is a vector with scalar values in the range $[0,1]$ with its shape determined by the distinct variations that we propose: 

\begin{itemize}
    \item \textbf{Global [G]: }One global value. 
    \item \textbf{Temporal [T]: }One value per frame in the sequence.
    \item \textbf{Spatial [S]: }One value per body joint of an individual. 
    \item \textbf{Spatio-Temporal [ST]: }One value per joint and frame.
\end{itemize}

These variations allow the \textit{Mixer} to have more flexibility in determining how to mix the motions. Unlike previous approaches \cite{related:interaction-1,in2in}, with a minor additional computation cost, the \textit{Mixer} can predict distinct mixing weights for disparate motions allowing for a dynamic and unique mixing conditioned on the particularities of each model output.

\subsection{Mixing}
\label{sec:method:mixing}

We base our mixing formula on DualMDM~\cite{in2in}. This model composition technique is based on DiffusionBlending~\cite{related:interaction-1}, where two different motions can be combined using a constant weight $w$ that specifies the importance of each model in the blending. DualMDM proposes to use a weight scheduler that defines a weight at every step of the denoising process $w_t$ to have more fine-grained control over the influence of each model at each denoising step, based on the findings of \cite{huang2023collaborative,wang2024analysisclassifierfreeguidanceweight}. This mixing formula is defined as: 
\begin{equation}
        x^m_t = x_t^a + w_t \cdot (x_t^b - x_t^a).
\end{equation}

However, the DualMDM approach involves the specification of experimentally found manual and constant weight schedulers. Although this approach yields superior outcomes compared to previous methods, it lacks the capability to adapt the mixing weights for different motions and conditions. This is a crucial aspect, as different motions generated by distinct models may necessitate unique weights depending on their specific characteristics. Instead, our \textit{Mixer} module can predict dynamic mixing weights, such that:
\begin{equation} \label{eq:mixer}
        x^m_t = x_t^a + Mixer(x_t^a, c^a, x_t^b, c^b, t) \cdot (x_t^b - x_t^a).
\end{equation}

\subsection{Adversarial Training}
\label{sec:method:discriminative}
\begin{figure}
    \centering
    \includegraphics[width=\linewidth]{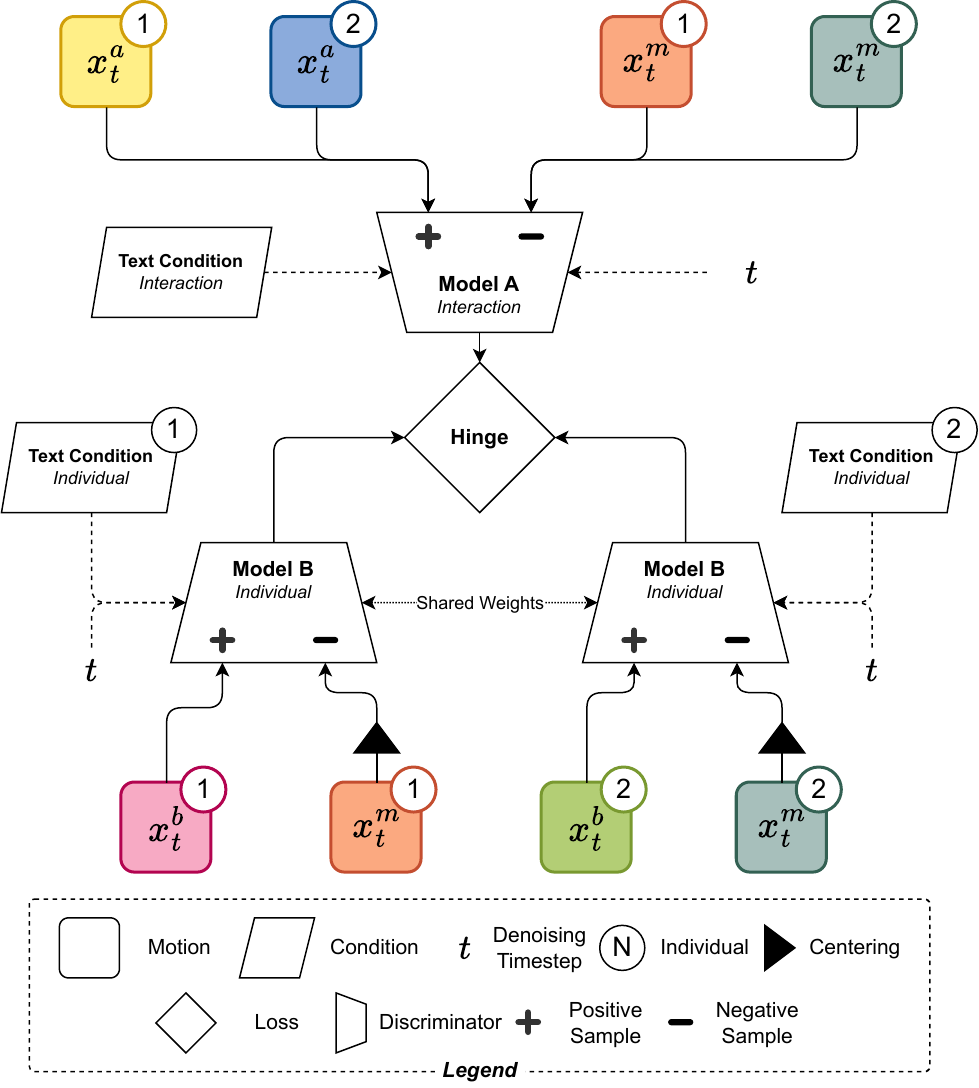}
    \caption{\textbf{Adversarial training.} Each pre-trained model has a specific discriminator that is trained with a hinge loss. We use the outputs of the pre-trained model as positive samples, and the mixed predictions generated by MixerMDM as negative samples.}
    \label{fig:discriminators}
    \vspace{-8mm}
\end{figure}

The resulting motion from MixerMDM is a unique mixing of two generated motions, and thus no ground truth can be employed to train the \textit{Mixer} in a supervised way. Inspired by \cite{goodfellow2014generative, huang2023collaborative}, we propose an adversarial training of the \textit{Mixer} with GANs (see \cref{fig:discriminators}). In this particular case, we have one discriminator per each pre-trained model. The training goal for the Mixer (generator) is to be able to generate a mixed motion that can fool both discriminators. These discriminators take the motions generated by the pre-trained models as positive samples (real), and the mixed motions generated by MixerMDM as negative ones (fake). In accordance with \cite{gan:diffusion2}, the hinge loss is employed as the adversarial objective function. The generator is trained to minimize: 

\begin{equation}
\mathcal{L}_{\mathrm{adv}}^{\mathrm G} = - \mathcal{D}^{a}(x^m_t) - \mathcal{D}^{b}(x^m_t) + L1,
\end{equation}

whereas the discriminator is trained to minimize:

\begin{equation}
\begin{split}
\mathcal{L}_{\mathrm{adv}}^{\mathrm D} = &\mathrm{min}(0,-1-\mathcal{D}^{a}(x^m_t)) + \mathrm{min}(0,-1-\mathcal{D}^{b}(x^m_t)) +\\
&\mathrm{min}(0,-1+\mathcal{D}^{a}(x^a_t)) + \mathrm{min}(0,-1+\mathcal{D}^{b}(x^b_t)) +\\
&L1,
\end{split}
\end{equation}

where $\mathcal{D}$ are the predictions of a given discriminator and L1 is a regularization loss that is applied to penalize big differences between the losses of the distinct discriminators. The generator and discriminators are trained interleaved.

Such adversarial training procedure allows training MixerMDM without having explicit ground truth motion, and preserving the core characteristics of each of the pre-trained models by fooling their respective discriminators.

\section{Experiments}
\label{sec:results}

\subsection{Datasets}
Our experiments employ the InterHuman \cite{data:intergen} and HumanML3D \cite{data:humanml3d} datasets. InterHuman stands as one of the most extensive annotated human-human interaction datasets available, containing 7,779 interactions labeled with textual descriptions. While including a diverse range of interactions, the intrapersonal diversity is limited. Conversely, HumanML3D is a dataset containing 14,616 individual motions. Although no interpersonal dynamics are included in this dataset, the intrapersonal diversity is far superior compared to InterHuman. The individual pre-trained models employed by MixerMDM are trained on HumanML3D, whereas the interaction counterparts are trained on InterHuman, both using textual descriptions. We further leverage InterHuman to train the \textit{Mixer}, using the individual textual descriptions generated by \cite{in2in}. These individual descriptions are used as conditioning for generating the motions from the individual models. For consistency across experiments, we adapted the HumanML3D motion representation to align with the one used in the InterHuman dataset.

\subsection{Evaluation Metrics} 
The metrics proposed in \cite{data:humanml3d} are usually employed to evaluate Human Motion Generation. Among them, we can find R-Precision, Multimodal-Dist, FID, Diversity, and Multimodality. Each of them is used to evaluate certain aspects of the motion generation. Specifically, R-Precision and Multimodal-Dist assess the semantic similarity between the generated motions and the input prompts. However, for the specific challenge addressed in this paper, these metrics fall short in reflecting the quality of the mixing process, which we define as the capability of a method to blend the predictions made by an interaction and individual model, maintaining the overall interaction while being precise in motions from the different individuals. To address this limitation, \cite{in2in} introduced the EID metric, which captures intrapersonal diversity within human-human interactions by randomly replacing individual motions. However, it does not consider the interaction and individual quality of the mixing. Therefore, we introduce:

\noindent\textbf{Alignment.} This metric uses the R-Precision (Top-3) to represent the alignment between the condition and the generated motion. However, instead of calculating this metric directly using the dataset used to train the Mixer,  we calculate the alignment of the mixed motions with respect to each dataset used by the pre-trained models in the pipeline. With this procedure, we quantify how much the blended motion preserves the characteristics of their base pre-trained models. In the particular case that we present, the Interaction and Individual \textit{Alignment} are calculated by computing the R-Precision of MixerMDM over the InterHuman and HumanML3D datasets, respectively. Two distinct feature extractors are used to calculate this metric on each dataset. Ultimately, an \textit{Overall Alignment} score is derived by computing the harmonic mean of these two alignment metrics, providing a more interpretable measure for identifying the optimal mixing model.

\noindent\textbf{Adaptability.} Unlike previous methods, MixerMDM can predict distinct mixing weights for different motions and conditions. \textit{Adaptability} indicates the model's capability to adapt the mixing of two motions depending on their particularities. It is computed as the average standard deviation of the mixing weights over the test set of InterHuman.

\subsection{Implementation Details}
The Mixer consists of four consecutive multi-head attention layers, each with a latent dimension of 512 and eight heads, resulting in a model with 21M parameters —significantly fewer than the over 300M parameters of the pre-trained models utilized. We employ a frozen CLIP-ViT$L/14$ model \cite{CLIP} as the text encoder, shared across all models in the pipeline. The number of diffusion timesteps is set to 1K, with a cosine noise schedule \cite{cosine}. During inference, we apply DDIM sampling \cite{related:diffusion-ddim} with $\eta=0$ and 50 timesteps. All models are trained using the AdamW optimizer~\cite{adamw}, with betas of $(0.9, 0.999)$, a weight decay of $10^{-5}$, and a learning rate of $10^{-5}$. Training incorporates adversarial losses as detailed in \cref{sec:method:discriminative} with an L1 weight of $0.1$. Each model undergoes 300 training epochs with a batch size of 128, utilizing gradient accumulation and 16-bit mixed precision. Training took 36 hours on a single Nvidia 4090 GPU.

\begin{figure*}[ht]
    \centering
    \includegraphics[width=\linewidth]{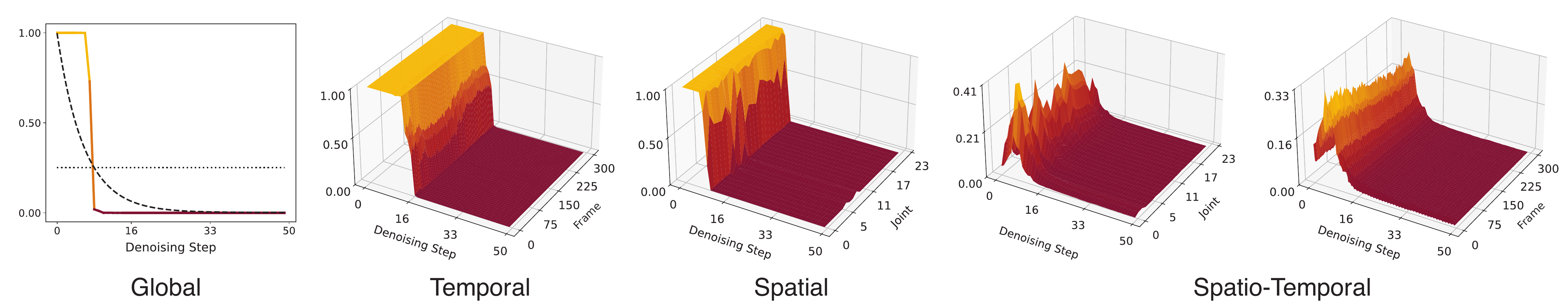}
    \caption{\textbf{Mean mixing weights.} The mean mixing weights of the best models for each variation of the \textit{Mixer} output. Previous model composition techniques appear in the \textit{Global} plot with a dotted line (DiffusionBlending~\cite{related:interaction-1}) and a dashed line (DualMDM~\cite{in2in}).}
    \label{fig:mean}
    \vspace{-3mm}
\end{figure*}

\subsection{Comparison to state-of-the-art Approaches} 
\label{sec:results:qualitative}
We compare MixerMDM against DiffusionBlending~\cite{related:interaction-1} and DualMDM~\cite{in2in} as previous methods for mixing human motion models. Additionally, a model trained with HumalML3D and finetuned with InterHuman is used as a baseline for the comparisons (\textit{Finetuned}). Different versions of MixerMDM are evaluated by incorporating the different mixing weights variations explained in \cref{sec:method:mixer} with different combinations of pre-trained models. We select InterGen~\cite{data:intergen} and in2IN~\cite{in2in} models for human-human motion generation ($\mathcal{M}^a$), and MDM~\cite{classes-3} and the individual version of in2IN for individual motion generation($\mathcal{M}^b$).

\noindent\textbf{Quantitative Comparisons.} \cref{tab:evaluations} presents the quantitative evaluations conducted across all MixerMDM variations. The results indicate that the version using Spatio-Temporal (ST) mixing weights, alongside the interaction and individual versions of in2IN, achieves the highest performance in terms of \textit{Overall Alignment}. Concerning \textit{Adaptability}, this model achieves the third-best score. However, the best models in this metric have a much lower \textit{Alignment}, suggesting that they sacrifice the mixing quality to achieve more adaptability. As the main goal of this method is to achieve the best blending possible, we consider the ST model with $\mathcal{M}^a=$ in2IN and $\mathcal{M}^b=$ in2IN as the best trade-off. With respect to the previous state-of-the-art approaches, almost all versions of MixerMDM outperform previous approaches. Additionally, the \textit{Adapatability} component is a new indicator that is not present in previous approaches. While higher \textit{Adaptability} does not mean higher quality, this value alongside better quantitative performance, reinforces the claim that MixerMDM benefits from our new learnable dynamic mixing strategy, compared to existing static ones.

\begin{table}[h]
\centering
\setlength{\tabcolsep}{3pt}
\footnotesize

\begin{tabular}{c:c:c|ccc}
\toprule
\multirow{2}{*}{Method} & \multirow{2}{*}{$\mathcal{M}^a$} & \multirow{2}{*}{$\mathcal{M}^b$} & \multicolumn{3}{c}{Alignment $\uparrow$} \\
\cline{4-6}
 & & & Interaction & individual & Overall  \\
\hline
Finetuned & - & - & $.675^{ \pm .01}$ & $.184^{ \pm .01}$ & $.289^{ \pm .00}$ \\
Diff.Blending~\cite{related:interaction-1}  & in2IN  & in2IN & $.577^{ \pm .00}$ & $.137^{ \pm .02}$ & $.221^{ \pm .01}$  \\
DualMDM~\cite{in2in} & in2IN  & in2IN & $.574^{ \pm .00}$ & $.134^{ \pm .01}$ & $.217^{ \pm .00}$ \\
\bottomrule
\end{tabular}
\begin{tabular}{c:c:c|ccc|c}
\toprule
\multirow{2}{*}{$\mathcal{M}^a$} & \multirow{2}{*}{$\mathcal{M}^b$} & \multirow{2}{*}{Type} & \multicolumn{3}{c|}{Alignment $\uparrow$} & \multirow{2}{*}{Adp.}\\
\cline{4-6}
 & & & Interaction & individual & Overall &  \\
\hline
\multirow{8}{*}{\parbox{0.95cm}{\centering in2IN\\\cite{in2in}}} & \multirow{4}{*}{\parbox{0.95cm}{\centering in2IN\\\cite{in2in}}} & G & $.521^{ \pm .00}$ & $.228^{ \pm .01}$ & $.317^{ \pm .00}$ & $.004$ \\
& & T & $.672^{ \pm .02}$ & $.150^{ \pm .02}$ & $.245^{ \pm .01}$ & $.002$ \\
& & S & $.391^{ \pm .01}$ & $.257^{ \pm .01}$ & $.310^{ \pm .00}$ & $.015$ \\
& & ST & $.406^{ \pm .01}$ & $.286^{ \pm .01}$ & $\mathbf{.335^{ \pm .01}}$ & $.112$ \\
\cdashline{2-7}[2pt/4pt]
& \multirow{4}{*}{\parbox{0.95cm}{\centering MDM\\\cite{classes-3}}} & G & $.458^{ \pm .02}$ & $.172^{ \pm .01}$ & $.250^{ \pm .01}$ & $.000$ \\
& & T & $.391^{ \pm .01}$ & $.203^{ \pm .01}$ & $.267^{ \pm .00}$ & $.059$ \\
& & S & $.505^{ \pm .02}$ & $.161^{ \pm .01}$ & $.244^{ \pm .01}$ & $\mathbf{.222}$ \\
& & ST & $.385^{ \pm .01}$ & $.230^{ \pm .01}$ & $.288^{ \pm .00}$ & $.107$ \\
\cdashline{1-7}[2pt/4pt]
\multirow{8}{*}{\parbox{0.95cm}{\centering InterGen\\\cite{data:intergen}}} & \multirow{4}{*}{\parbox{0.95cm}{\centering in2IN\\\cite{in2in}}} & G & $.380^{ \pm .01}$ & $.272^{ \pm .01}$ & $.317^{ \pm .00}$ & $.004$ \\
& & T & $.328^{ \pm .01}$ & $.297^{ \pm .01}$ & $.312^{ \pm .00}$ & $.020$ \\
& & S & $.266^{ \pm .01}$ & $.281^{ \pm .01}$ & $.273^{ \pm .01}$ & $.004$ \\
& & ST & $.276^{ \pm .01}$ & $.313^{ \pm .01}$ & $.293^{ \pm .00}$ & $.013$ \\
\cdashline{2-7}[2pt/4pt]
& \multirow{4}{*}{\parbox{0.95cm}{\centering MDM\\\cite{classes-3}}} & G & $.416^{ \pm .01}$ & $.180^{ \pm .01}$ & $.251^{ \pm .00}$ & $.001$ \\
& & S & $.438^{ \pm .01}$ & $.181^{ \pm .01}$ & $.256^{ \pm .00}$ & $.022$ \\
& & T & $.490^{ \pm .02}$ & $.196^{ \pm .01}$ & $.280^{ \pm .01}$ & $.125$ \\
& & ST & $.489^{ \pm .02}$ & $.172^{ \pm .02}$ & $.254^{ \pm .02}$ & $.030$ \\
\bottomrule
\end{tabular}
\caption{\textbf{Quantitative evaluation}. Top: state-of-the-art comparison. Bottom: ablation of all the variations tested with MixerMDM. \textit{Type}: type of mixing weights predicted by the \textit{Mixer}. \textit{Adp.}: Adaptability metric. All evaluations are executed 10 times to elude the randomness of the generation. $\pm$ indicates the 95\% confidence interval. \textbf{Best} results are highlighted.}
\label{tab:evaluations}
\vspace{-3mm}
\end{table}

\cref{fig:mean} illustrates the mean mixing weights for the top-performing models. The higher the curve, the more importance is given to the individual model with respect to the interaction. These plots reveal patterns that, while bearing some resemblance to those of DualMDM, diverge considerably, particularly in the ST version, where the overall curve differs markedly. The common factor of all the learned curves is that the individual model has more importance at the beginning of the denoising process, while the interaction takes all the importance on the last steps of the diffusion process, validating the hypothesis stated in \cite{in2in}. 

\begin{figure}[h]
    \centering
    \includegraphics[width=\linewidth]{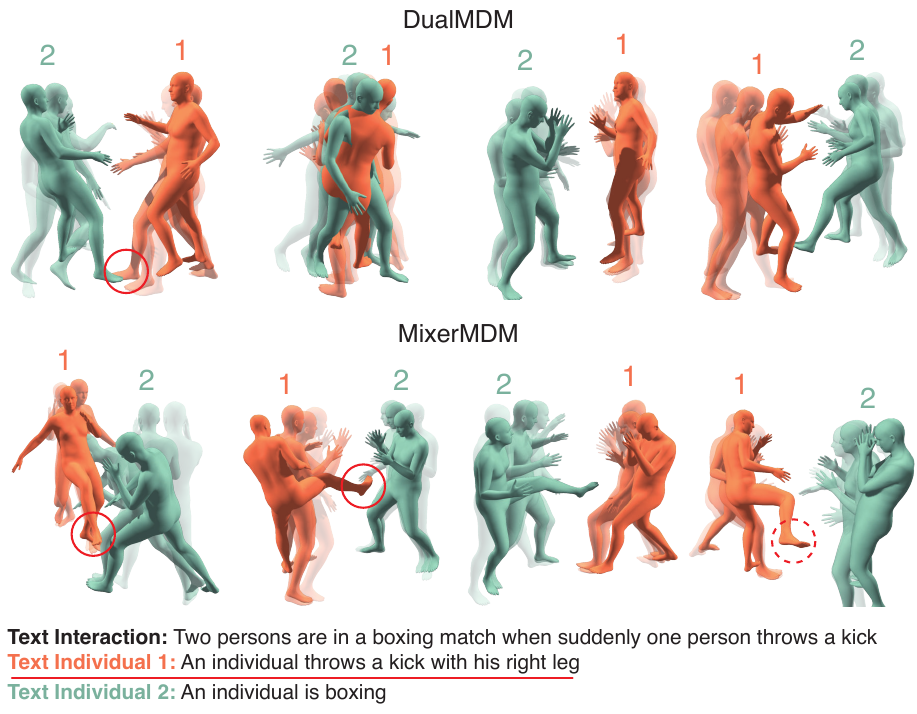}
    \caption{\textbf{Consistency.} When an individual variation (underline) is performed in one of the interactions,  MixerMDM achieves a greater level of consistency and individual assignment generating the mixed motion.}
    \label{fig:consistency1}
    \vspace{-3mm}
\end{figure}

\begin{figure*}[h]
  \centering
  \includegraphics[width=0.84\textwidth]{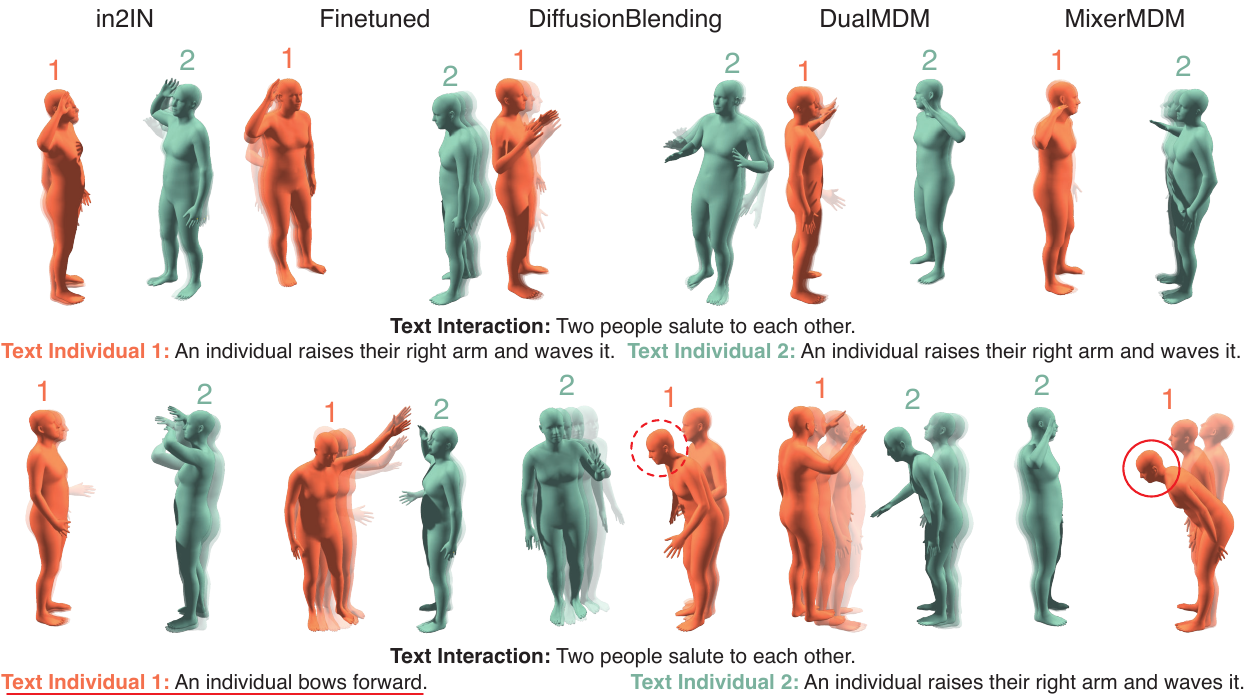}
  \caption{\textbf{Controllability}. While all methods can properly generate an interaction (top), when a variation in one of the individual conditions is applied (bottom, underline), MixerMDM generates the most aligned motion to the overall interaction and individual textual descriptions. }
  \label{fig:qualitative2}
    \vspace{-3mm}
\end{figure*}

\noindent\textbf{Qualitative Comparisons.} Beyond quantitative evaluation, qualitative comparisons demonstrate subtle differences between MixerMDM and previous methods.  \cref{fig:consistency1} shows how MixerMDM is much more consistent in generating mixed motions aligned with their conditioning than previous approaches. Additionally, in \cref{fig:qualitative2} we can observe that MixerMDM can generate more fine-grained individual variations to interaction from the InterHuman test set. For additional examples and interactive visualizations, please refer to the supplementary video.

Finally, we have conducted a user study to provide a more comprehensive evaluation of the method’s performance. 35 users ranked the evaluated methods based on the alignment of 20 randomly selected generated motions with 1) their interaction and 2) their individual textual descriptions. \cref{tab:userstudy} shows that MixerMDM significantly outperforms previous methods.

\begin{table}[H]
\centering
\setlength{\tabcolsep}{3pt}
\footnotesize
\begin{tabular}{c|cc|cc}
\toprule
 \multirow{2}{*}{Method} & \multicolumn{2}{c|}{Interaction} & \multicolumn{2}{c}{individual} \\
\cline{2-5}
 & Avg. $\downarrow$ & 1st $\uparrow$ & Avg. $\downarrow$ & 1st $\uparrow$ \\
\hline
Diff.Blending~\cite{related:interaction-1} & $2.531^{\pm .584}$ & $4.57\%$ & $2.446 ^{\pm .678}$ & $10.57\%$ \\
DualMDM~\cite{in2in} & $2.286^{\pm .641}$ & $10.29\%$ & $2.051^{\pm .736}$ & $24.57\%$ \\
MixerMDM (ours) & $\mathbf{1.182^{\pm .467}}$ & $\mathbf{85.14\%}$ & $\mathbf{1.309^{\pm .573}}$ & $\mathbf{74.86\%}$ \\
\bottomrule
\end{tabular}
\vspace{-1.5mm}
\caption{\textbf{User study}. Average rank (Avg), First ranked (1st). $\pm$ indicates standard deviation. }
\label{tab:userstudy}
\vspace{-3mm}
\end{table}

\subsection{Modularity}
One advantage of MixerMDM is its ability to mix motions by directly using the outputs of pre-trained models. This allows the \textit{Mixer} to learn how to mix motion distributions rather than focusing on model-specific features. This approach enables the reuse of weights of a trained \textit{Mixer} with new specialized models that can generate a similar distribution of motions to the ones originally used to train the \textit{Mixer}. In other words, pre-trained models can be substituted by others that have been trained with the same specific dataset. This provides a modular capability to the method, which is especially useful in a fast-paced field such as human motion generation. We evaluate this property in \cref{tab:swap}, where models with initially lower performance are combined with Mixer weights from the top-performing models. This leads to a notable performance boost without requiring any additional training.

\begin{table}[ht]
\centering
\setlength{\tabcolsep}{1.5pt}
\footnotesize
\begin{tabular}{c:c:c|ccc|c|c}
\toprule
\multirow{2}{*}{$\mathcal{M}^a$} & \multirow{2}{*}{$\mathcal{M}^b$} & \multirow{2}{*}{Type} & \multicolumn{3}{c|}{Alignment$\uparrow$} & \multirow{2}{*}{Adp.} & \multirow{2}{*}{Impr.$\uparrow$} \\
\cline{4-6}
 & & & Interaction & individual & Overall & & \\
\hline
 \multirow{4}{*}{\parbox{1cm}{\centering InterGen\\\cite{data:intergen}}} & \multirow{4}{*}{\parbox{1cm}{\centering MDM\\\cite{classes-3}}} & G & $.385^{ \pm .02}$ & $.313^{ \pm .01}$ & $\mathbf{.345^{ \pm .00}}$ & $.004$ & $\mathbf{37\%}$\\
 & & T & $.589^{ \pm .01}$ & $.232^{ \pm .01}$ & $.333^{ \pm .00}$ & $.001$ & $30\%$ \\
 & & S & $.380^{ \pm .01}$ & $.315^{ \pm .01}$ & $.344^{ \pm .00}$ & $.014$ & $23\%$ \\
 & & ST & $.417^{ \pm .01}$ & $.292^{ \pm .01}$ & $.344^{ \pm .00}$ & $\mathbf{.126}$ & $35\%$ \\
\bottomrule
\end{tabular}
\caption{\textbf{Modularity Evaluation}. The worst combination of pre-trained models is evaluated by using the weights of the best Mixer from \cref{tab:evaluations}. \textit{Impr.}: relative \textit{Overall Alignment} improvement (\%) with respect to the original evaluation.}
\label{tab:swap}
    \vspace{-3mm}
\end{table}
\section{Conclusion}
\label{sec:conclusion}
We introduced MixerMDM, a learnable approach for combining motions from distinct text-conditioned human motion diffusion models. Unlike previous methods, which rely on manually set parameters for mixing, MixerMDM learns unique mixing weights depending on the specific motions to combine through adversarial losses. Mixing models that generate individual motions with models that generate human-human interaction has resulted in MixerMDM consistently producing blended motions with a higher degree of individual controllability with respect to previous methods. Additionally, as MixerMDM learns to blend motions by directly using the outputs of the pre-trained models, it has a modular property that allows the swapping of pre-trained models without the need for re-training. Finally, we introduced an evaluation pipeline to quantitatively asses all these capabilities. 

\noindent\textbf{Limitations and Future Work. } Although MixerMDM is approximately ten times smaller than the pre-trained models it integrates, it still incurs additional computational costs and extends inference time, as mixing weights must be calculated at each denoising step. Moreover, due to the nature of adversarial training, achieving stability and tuning hyperparameters can be challenging. A further limitation and direction for future work lies in data representation discrepancies. For effective model mixing, models must predict a unified data representation, which is not always achievable directly and sometimes necessitates retraining models to ensure representation compatibility.

\clearpage
\textbf{Acknowledgments}
This publication is part of the TSI-100927-2023-6 Project, funded by the Recovery, Transformation and Resilience Plan from the European Union Next Generation through the Ministry for Digital Transformation and the Civil Service. This work has been partially supported by the Spanish project PID2022-136436NB-I00, the Spanish national grant for PhD studies, FPU22/04200, by ICREA under the ICREA Academia programme, and CIAICO/2022/132 Consolidated group project “AI4Health” funded by Valencian government.
{
    \small
    \bibliographystyle{ieeenat_fullname}
    \bibliography{main}
}

\renewcommand*{\thesection}{\Alph{section}}
\renewcommand*{\thefigure}{\Alph{figure}}
\renewcommand*{\thetable}{\Alph{table}}
\setcounter{section}{0}
\setcounter{table}{0}
\setcounter{figure}{0}
\clearpage
\setcounter{page}{1}
\maketitlesupplementary

This supplementary material aims to enhance the reproducibility and understanding of the previously presented contributions. In \cref{supp:datasets}, we detail the datasets used and describe the modifications made to integrate them into the MixerMDM pipeline. In \cref{supp:implementation}, we outline the implementation specifics of the state-of-the-art models used for comparison, as well as the evaluators employed in the proposed evaluation pipeline. In \cref{supp:quantitative}, we complement the quantitative evaluation with additional experiments and ablations. Lastly, in \cref{supp:qualitative}, we include additional visual examples illustrating the MixerMDM capabilities.

\section{Datasets}
\label{supp:datasets}

\subsection{InterHuman}
InterHuman~\cite{data:intergen} is one of the most extensive annotated datasets for human-human interactions, containing 7,779 interactions labeled with textual descriptions. Each individual's motion within an interaction is represented as a set of poses $x_{i}{=}\left[j^{p}, j^{v}, j^{r}, c^{f}\right]$, where $x^{i}$ denotes the $i$-th motion timestep. This representation includes joint positions and velocities $j^{p},j^{v} \,{\in}\, \mathbb{R}^{3 N_{j}}$ in the world frame, a 6D representation of local rotations $j^{r} \,{\in}\, \mathbb{R}^{6 N_{j}}$ in the root frame, and binary foot-ground contact features $c^{f} \,{\in}\, \mathbb{R}^{4}$. The number of joints in the InterHuman dataset is $N_j=22$. Each interaction in the dataset is paired with three textual descriptions summarizing the overall interaction. Additionally, the in2IN~\cite{in2in} framework introduced more detailed textual descriptions, generated by Large Language Models,  for the motions performed by each individual in the interaction. We utilized these detailed descriptions to condition the generation of individual models employed in the mixing process.

\subsection{HumanML3D}
The HumanML3D~\cite{data:humanml3d} dataset contains 14,616 individual motions annotated with textual descriptions. Each motion is represented as a set of poses $x_{i}{=}\left[\dot{r}^a, \dot{r}^x, \dot{r}^z,r^y, j^{p}, j^{v}, j^{r}, c^{f}\right]$, where $x^{i}$ denotes the $i$-th motion timestep. In this format, $\dot{r}^a \in \mathbb{R}$ is the root angular velocity in the Y-axis, $\dot{r}^x, \dot{r}^z \in \mathbb{R}$ are the root linear velocities on the XZ-plane, $r^y \in \mathbb{R}$ is the root height, $j^{p},j^{v} \,{\in}\, \mathbb{R}^{3 N_{j}}$ and $j^{r} \,{\in}\, \mathbb{R}^{6 N_{j}}$ are the joint positions, velocities, and rotations in the root frame, and $c^{f} \,{\in}\, \mathbb{R}^{4}$ are binary foot-ground contact features. The number of joints in the HumanML3D dataset is $N_j=22$. Since this representation differs from the format used in InterHuman and is not optimal for capturing the relative positions of interactants, we have converted it to the InterHuman format. This conversion involves processing the raw SMPL motions from HumanML3D to extract the global joint positions and velocities as well as the relative rotations as it is done in the InterHuman pre-processing.

\section{Further Implementation Details}
\label{supp:implementation}

\subsection{Motion Transformations}
\label{supp:transformations}
Motion transformations are applied to maintain the pre-trained models within their learned distribution. The \textit{centering} function translates the motion to the origin of coordinates in the XZ plane, simultaneously orienting the trajectory initially in the Z+ direction. The \textit{alignment} is a global transformation applied to a motion ($x^a$) with respect to another ($x^b$). Firstly, $x^a$ is translated to the initial position of $x^b$. Secondly, $x^a$ is rotated to match the orientation of the vector of the initial and end positions of $x^b$. The same transformation is applied to the whole motion, thus not introducing foot sliding, and standardizes global positioning and orientation to the individual models.

\subsection{State-of-the-art Implementations}
The methods employed in Sec. 4 were implemented using their respective official codebases. We leveraged the original checkpoints of InterGen~\cite{data:intergen} and the interaction and individual versions of in2IN~\cite{in2in}, as they were trained with the same motion representation we use. For MDM~\cite{classes-3}, originally trained on HumanML3D, we kept the architecture as is, and adapted the output shape of the denoiser to match the size of our motion representation.

\begin{table*}[ht]
\centering
\footnotesize
\setlength{\tabcolsep}{3pt}
\begin{tabular}{c:c|cc|cc|cc|cc|cc}
\toprule
\multirow{2}{*}{Method} & \multirow{2}{*}{Type} & \multicolumn{2}{c|}{R-Precision $\uparrow$} & \multicolumn{2}{c|}{FID $\downarrow$} & \multicolumn{2}{c|}{MM Dist $\downarrow$} & \multicolumn{2}{c|}{Diversity $\rightarrow$} & \multicolumn{2}{c}{MModality $\uparrow$} \\
\cline{3-12}
 & & Interaction & Individual & Interaction & Individual & Interaction & Individual & Interaction & Individual & Interaction & Individual \\
\hline
Ground Truth & - & $.701^{\pm .01}$ & $.563^{\pm .00}$ & $.273^{\pm .01}$ & $1.04^{\pm .14}$ & $3.76^{\pm .01}$ & $3.44^{\pm .00}$ & $7.95^{\pm .06}$ & $16.3^{\pm .05}$ & - & - \\
\hline
Diff.Blending~\cite{related:interaction-1} & - & $.577^{\pm .00}$ & $.137^{\pm .02}$ & $33.8^{\pm .29}$ & $360^{\pm 16}$ & $3.89^{\pm .00}$ & $5.18^{\pm .01}$ & $6.14^{\pm .14}$ & $11.9^{\pm .22}$ & $.779^{\pm .12}$ & $2.74^{\pm .01}$ \\
DualMDM~\cite{in2in} & - & $.574^{\pm .00}$ & $.134^{\pm .01}$ & $22.9^{\pm .19}$ & $330^{\pm .02}$ & $3.85^{\pm .01}$ & $5.13^{\pm .01}$ & $7.04^{\pm .17}$ & $12.5^{\pm .28}$ & $.935^{\pm .12}$ & $2.74^{\pm .09}$ \\
\hline
\multirow{4}{*}{\parbox{2cm}{\centering~MixerMDM\\(ours)}} 
 & G & $.521^{\pm .00}$ & $.228^{\pm .01}$ & $44.5^{\pm .99}$ & $199^{\pm 19}$ & $3.92^{\pm .00}$ & $4.70^{\pm .14}$ & $6.57^{\pm .19}$ & $14.0^{\pm .53}$ & $1.08^{\pm .19}$ & $2.99^{\pm .12}$ \\
 & T & $\mathbf{.672^{\pm .02}}$ & $.150^{\pm .02}$ & $\mathbf{21.2^{\pm .70}}$ & $245^{\pm 5.1}$ & $\mathbf{3.85^{\pm .00}}$ & $5.05^{\pm .00}$ & $\mathbf{7.57^{\pm .07}}$ & $13.6^{\pm .16}$ & $1.14^{\pm .25}$ & $3.04^{\pm .21}$ \\
 & S & $.391^{\pm .01}$ & $.257^{\pm .01}$ & $52.4^{\pm 1.8}$ & $192^{\pm 5.9}$ & $3.94^{\pm .00}$ & $4.69^{\pm .03}$ & $6.40^{\pm .20}$ & $14.4^{\pm .09}$ & $1.11^{\pm .02}$ & $2.96^{\pm .07}$ \\
 & ST & $.406^{\pm .01}$ & $\mathbf{.286^{\pm .01}}$ & $47.6^{\pm .88}$ & $\mathbf{142^{\pm .75}}$ & $3.93^{\pm .01}$ & $\mathbf{4.60^{\pm .02}}$ & $6.57^{\pm .09}$ & $\mathbf{15.1^{\pm .18}}$ & $\mathbf{1.23^{\pm .02}}$ & $\mathbf{3.26^{\pm .07}}$ \\
\bottomrule
\end{tabular}
\caption{\textbf{Quantitative evaluation with conventional metrics}. Ours \{G,T,S,TS\} uses $\mathcal{M}^a {=}\mathcal{M}^b{=}$ in2IN . Mean of 10 evaluations, $\pm$ shows the 95\% confidence interval. Best in \textbf{bold}.}
\label{tab:metrics}
\end{table*}

\subsection{Evaluators}
The evaluation metrics for human motion generation require a feature extractor that produces aligned latent representations of both the generated motions and the corresponding conditions (text, in this case). The feature extractor architecture is based on the one used in the InterHuman dataset: a \textit{MotionEncoder} and a \textit{TextEncoder}. The \textit{MotionEncoder} is a Transformer encoder with 8 layers of 8 heads each, which transforms the motion into a 2048-dimensional latent vector. This vector is then compressed to a dimension of 512 using a Multi-Layer Perceptron. The \textit{TextEncoder} is a frozen CLIP-ViT$L/14$ model \cite{CLIP}, supplemented with a Transformer encoder with 8 layers of 8 heads each to adapt the CLIP latent space to better match the dataset distribution. We trained a feature extractor for each dataset employed. These models were trained for 500 epochs with a batch size of 64, using the AdamW optimizer~\cite{adamw} with $\beta$ parameters set to $(0.9, 0.999)$, a weight decay of $10^{-5}$, and a learning rate of $10^{-4}$.

\section{Further Quantitative Examples}
\label{supp:quantitative}
In this section, we complement the quantitative study from \cref{sec:results:qualitative} with additional experiments and ablations. In addition to the proposed metrics, we have evaluated MixerMDM with conventional metrics. \cref{tab:metrics} shows that our method surpasses previous methods on the proposed and conventional metrics.

\subsection{Motion Transformations}
The \textit{centering} and \textit{alignment} transformation (\cref{sec:method:pipeline}, \cref{supp:transformations}) help to maintain the mixed motion within the distribution of the pre-trained models. \cref{tab:align} shows the effect of not using the alignment transformation. While producing a performance drop in the interaction evaluation, results still outperform previous methods.

\begin{table}[H]
\centering
\setlength{\tabcolsep}{2.6pt}
\scriptsize
\begin{tabular}{c|c|c|c|c|c}
\toprule
\multirow{2}{*}{Method} & \multicolumn{1}{c|}{Top-3 R-Prec. $\uparrow$} & \multicolumn{1}{c|}{FID $\downarrow$} & \multicolumn{1}{c|}{MM Dist $\downarrow$} & \multicolumn{1}{c|}{Diversity $\rightarrow$} & \multicolumn{1}{c}{MModality $\uparrow$} \\
\cline{2-6}
 & Interaction & Interaction & Interaction & Interaction & Interaction \\
\hline
G & $.391^{\pm .01}$ & $45.7^{\pm .31}$ & $3.92^{\pm .01}$ & $6.51^{\pm .08}$ & $\mathbf{1.15^{\pm .02}}$ \\
T & $\mathbf{.578^{\pm .00}}$ & $\mathbf{20.2^{\pm .04}}$ & $\mathbf{3.84^{\pm .00}}$ & $\mathbf{7.73^{\pm .09}}$ & $.990^{\pm .01}$ \\
S & $.375^{\pm .01}$ & $50.6^{\pm 1.0}$ & $3.93^{\pm .01}$ & $6.43^{\pm .01}$ & $1.01^{\pm .15}$ \\
ST & $.380^{\pm .02}$ & $41.1^{\pm .00}$ & $3.93^{\pm .01}$ & $6.62^{\pm .05}$ & $1.01^{\pm .01}$ \\
\bottomrule
\end{tabular}
\caption{\textbf{MixerMDM without alignment}. Compare with \cref{tab:metrics}.}
\label{tab:align}
\end{table}

\subsection{Usabilitiy}
While using more prompts allows higher controllability, it can hinder usability with tedious descriptions in cases where such controllability is not a priority. Using an LLM (gpt4o-mini) at inference allows using just the interaction prompt and inferring the individual ones. \cref{tab:llm} shows that this strategy does not affect the interaction motion quality and text-alignment.

\begin{table}[H]
\centering
\setlength{\tabcolsep}{2.6pt}
\scriptsize
\begin{tabular}{c|c|c|c|c|c}
\toprule
\multirow{2}{*}{Method} & \multicolumn{1}{c|}{Top-3 R-Prec. $\uparrow$} & \multicolumn{1}{c|}{FID $\downarrow$} & \multicolumn{1}{c|}{MM Dist $\downarrow$} & \multicolumn{1}{c|}{Diversity $\rightarrow$} & \multicolumn{1}{c}{MModality $\uparrow$} \\
\cline{2-6}
 & Interaction & Interaction & Interaction & Interaction & Interaction \\
\hline
G & $.451^{\pm .00}$ & $46.7^{\pm .18}$ & $3.92^{\pm .01}$ & $6.61^{\pm .12}$ & $1.03^{\pm .18}$ \\
T & $\mathbf{.651^{\pm .02}}$ & $\mathbf{21.2^{\pm .77}}$ & $\mathbf{3.83^{\pm .01}}$ & $\mathbf{7.69^{\pm .17}}$ & $.998^{\pm .00}$ \\
S & $.412^{\pm .03}$ & $49.3^{\pm 1.2}$ & $3.93^{\pm .01}$ & $6.40^{\pm .09}$ & $\mathbf{1.11^{\pm .00}}$ \\
ST & $.341^{\pm .00}$ & $49.1^{\pm 1.4}$ & $3.94^{\pm .00}$ & $6.39^{\pm .04}$ & $1.03^{\pm .05}$ \\
\bottomrule
\end{tabular}
\caption{\textbf{MixerMDM LLM aided}. Compare with \cref{tab:metrics}.}
\label{tab:llm}
\end{table}

\section{Further Qualitative Examples}
\label{supp:qualitative}
\begin{figure}[h]
    \centering
    \includegraphics[width=\linewidth]{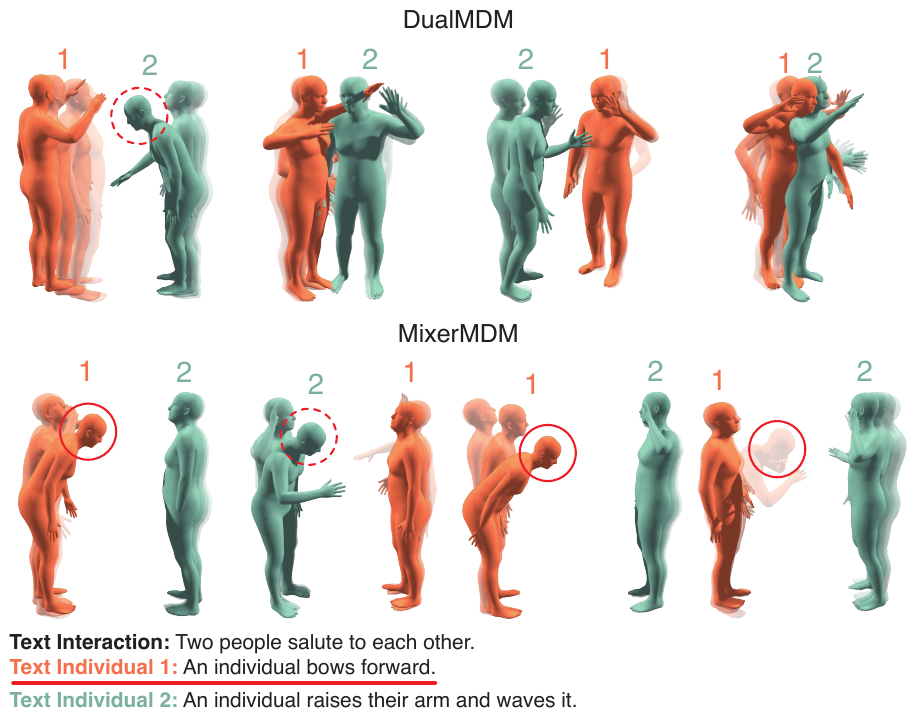}
    \caption{\textbf{Consistency.} When an individual variation (underlined) is performed in one of the interactions,  MixerMDM achieves a greater level of consistency generating the mixed motion.}
    \label{fig:supp:qualitative3}
\end{figure}

\begin{figure*}[t]
  \centering
  \begin{subfigure}{0.95\textwidth}
    \centering
    \includegraphics[width=\textwidth]{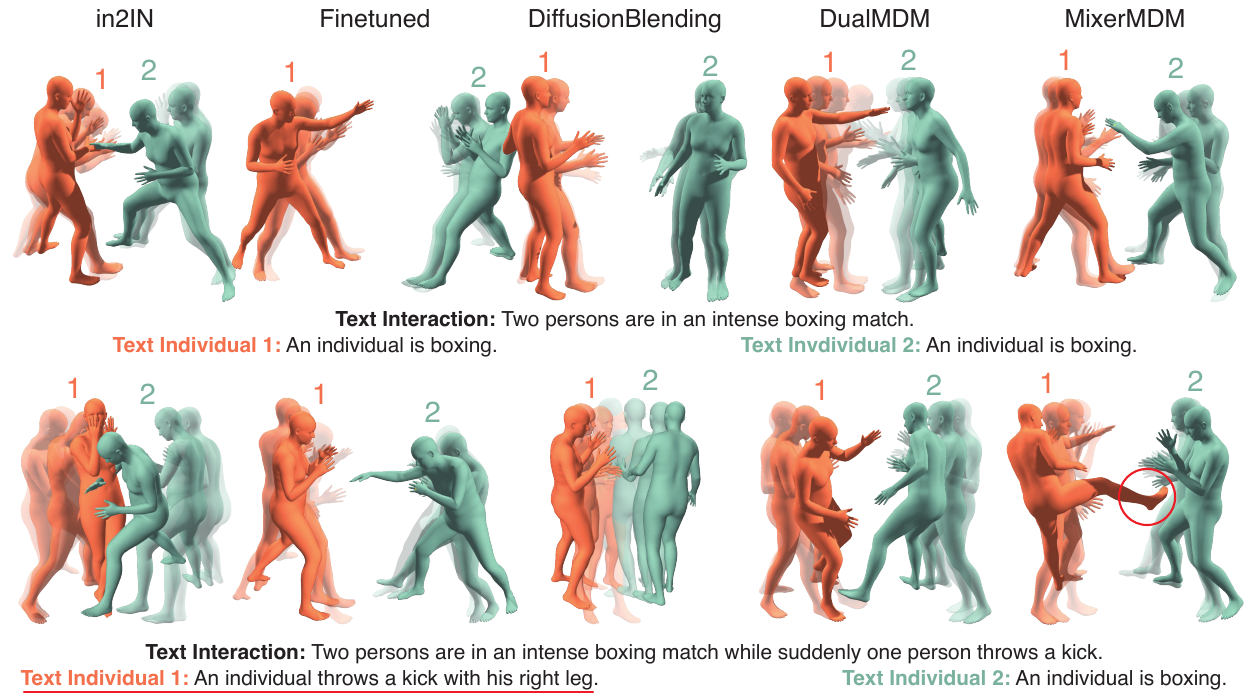}
    \caption{Boxing interaction}
    \label{fig:supp:qualitative2a}
  \end{subfigure}
  \vspace{2mm}
  \vspace{0.25cm}
  \begin{subfigure}{0.95\textwidth} 
    \centering
    \includegraphics[width=\textwidth]{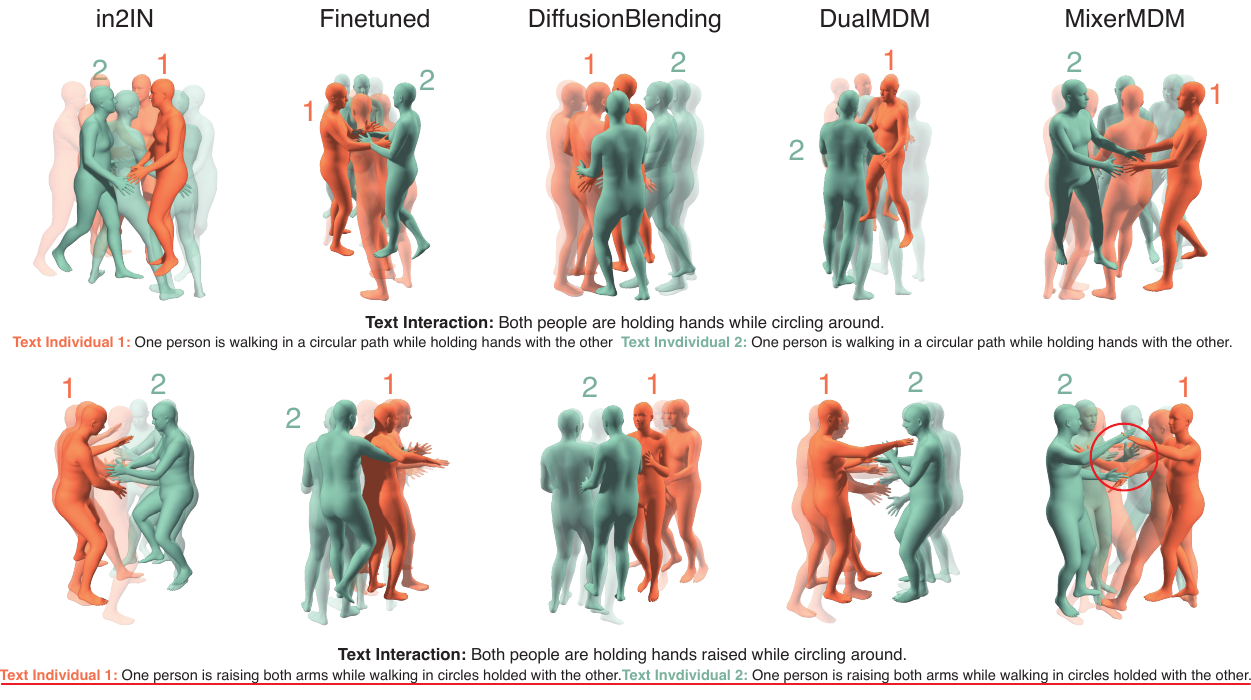}
    \caption{Walking in circles interaction}
    \label{fig:supp:qualitative2b}
  \end{subfigure}
  \caption{\textbf{Controllability}. While all methods can properly generate an interaction (top), when a variation in one of the individual conditions is applied (bottom, underlined), MixerMDM generates the most aligned motion to the overall interaction and individual textual descriptions.}
  \label{fig:supp:qualitative2}
\end{figure*}

In this section, we complement the qualitative study from \cref{sec:results:qualitative} with additional examples. \cref{fig:supp:qualitative2} show the superior individual controllability of MixerMDM when compared to previous approaches. With MixerMDM, we can achieve detailed control of the individual motions while still preserving the dynamics of the interaction. This is achieved thanks to the adversarial training that promotes preserving the main interaction dynamics as well as the individual ones. \cref{fig:supp:qualitative3} shows another example of consistency on this dual control. We refer the reader to the attached video for a more detailed visualization of all the examples that we showed and discussed in this section.

\end{document}